\def\year{2020}\relax
\documentclass[letterpaper]{article} % DO NOT CHANGE THIS
\usepackage{aaai20}  % DO NOT CHANGE THIS
\usepackage{times}  % DO NOT CHANGE THIS
\usepackage{helvet} % DO NOT CHANGE THIS
\usepackage{courier}  % DO NOT CHANGE THIS
\usepackage[hyphens]{url}  % DO NOT CHANGE THIS
\usepackage{graphicx} % DO NOT CHANGE THIS
\usepackage{graphicx}  % DO NOT CHANGE THIS
\usepackage{amsmath}
\usepackage{amssymb}
\usepackage{hyperref}
\usepackage{url}
\usepackage{subcaption}
\usepackage[linesnumbered,ruled,vlined,english,onelanguage]{algorithm2e}
\usepackage{multirow}
\usepackage{blindtext}
\newcommand{\citet}[1]{\citeauthor{#1} \shortcite{#1}}
\usepackage{xcolor}

\title{Data Efficient Direct Speech-to-Text Translation with Modality Agnostic Meta-Learning}
\author{Sathish Indurthi, Houjeung Han, Nikhil Kumar Lakumarapu, \\ \Large \textbf{Beomseok Lee, Insoo Chung, Sangha Kim, Chanwoo Kim} \\
Samsung Research, Seoul, South Korea \\
\{s.indurthi, h.j.han, n07.kumar, bsgunn.lee, sangha01.kim, insooo.chung, chanw.com\}@samsung.com}

\begin{document}

\maketitle
\begin{abstract}
End-to-end Speech Translation (ST) models have several advantages such as lower latency, smaller model size, and less error compounding over conventional pipelines that combine Automatic Speech Recognition (ASR) and text Machine Translation (MT) models. However, collecting large amounts of parallel data for ST task is more difficult compared to the ASR and MT tasks. Previous studies have proposed the use of transfer learning approaches to overcome the above difficulty. These approaches benefit from weakly supervised training data, such as ASR speech-to-transcript or MT text-to-text translation pairs.  However, the parameters in these models are updated independently of each task, which may lead to sub-optimal solutions. In this work, we adopt a meta-learning algorithm to train a modality agnostic multi-task model that transfers knowledge from source tasks=ASR+MT  to target task=ST where ST task severely lacks data. In the meta-learning phase, the parameters of the model are exposed to vast amounts of speech transcripts (e.g., English ASR) and text translations (e.g., English-German MT). During this phase, parameters are updated in such a way to understand speech, text representations,  the relation between them, as well as act as a good initialization point for the target ST task. We evaluate the proposed meta-learning approach for ST tasks on English-German (En-De) and English-French (En-Fr) language pairs from the Multilingual Speech Translation Corpus (MuST-C). Our method outperforms the previous transfer learning approaches and sets new state-of-the-art results for En-De and En-Fr ST tasks by obtaining 9.18, and 11.76 BLEU point improvements, respectively.
\end{abstract}

\section{Introduction}
\label{sec:intro}
 The Speech Translation (ST) task takes audio as input and generates text translation as output. Traditionally it is achieved by cascading Automatic Speech Recognition (ASR) and Machine Translation (MT) models \cite{ney1999speech}. However, the cascaded model suffers from compounding errors between ASR and MT models, higher latency due to sequential inference from the two models, and higher memory and computational resource requirements.
 
 End-to-end neural models for Automatic Speech Recognition (ASR) \cite{ASR} and Machine Translation (MT) \cite{Bahdanau14} are evolving into end-to-end neural model for Speech Translation (ST)\cite{berard2016listen}. Such models overcome the above limitations of cascaded systems. However, training such end-to-end ST models requires huge amounts of speech-to-text parallel data. Huge amounts of parallel data along with the advancements in sequence-to-sequence models led to successful ASR and MT neural systems. However, collecting such amounts of parallel data for training ST system is very challenging. 
 
 To alleviate the issue of collecting vast amounts of parallel data for ST task,  \citet{berard2018end,jia2019leveraging} proposed pre-training based approaches such as transfer learning.  Although these approaches have been shown to improve the performance of the ST task, they have some limitations. 
 In the transfer learning strategy, the pre-trained model parameter updates are based on the current task at hand and are not optimized towards adapting to new tasks. In multi-task learning \cite{weiss2017sequence,anastasopoulos2018tied}, a variant of transfer learning, the parameters are shared across multiple tasks, and thus limits the performance of individual tasks. Moreover, the parameters of the model are updated independently based on individual task performance, and this may lead to sub-optimal solutions in these approaches.
 %In transfer learning approach and its variant multi-task learning \cite{weiss2017sequence,anastasopoulos2018tied}, the parameters of the model are updated independently based on individual task performance, and this may lead to a sub-optimal solution in both approaches. Moreover, in multi-task learning, the parameters are shared across multiple tasks, which limits the performance of individual tasks.
 
  \begin{figure*}[t]
   \centering
    \hspace{-2cm}
    \begin{subfigure}[t]{0.6\textwidth}
       \centering
        \includegraphics[height=6cm]{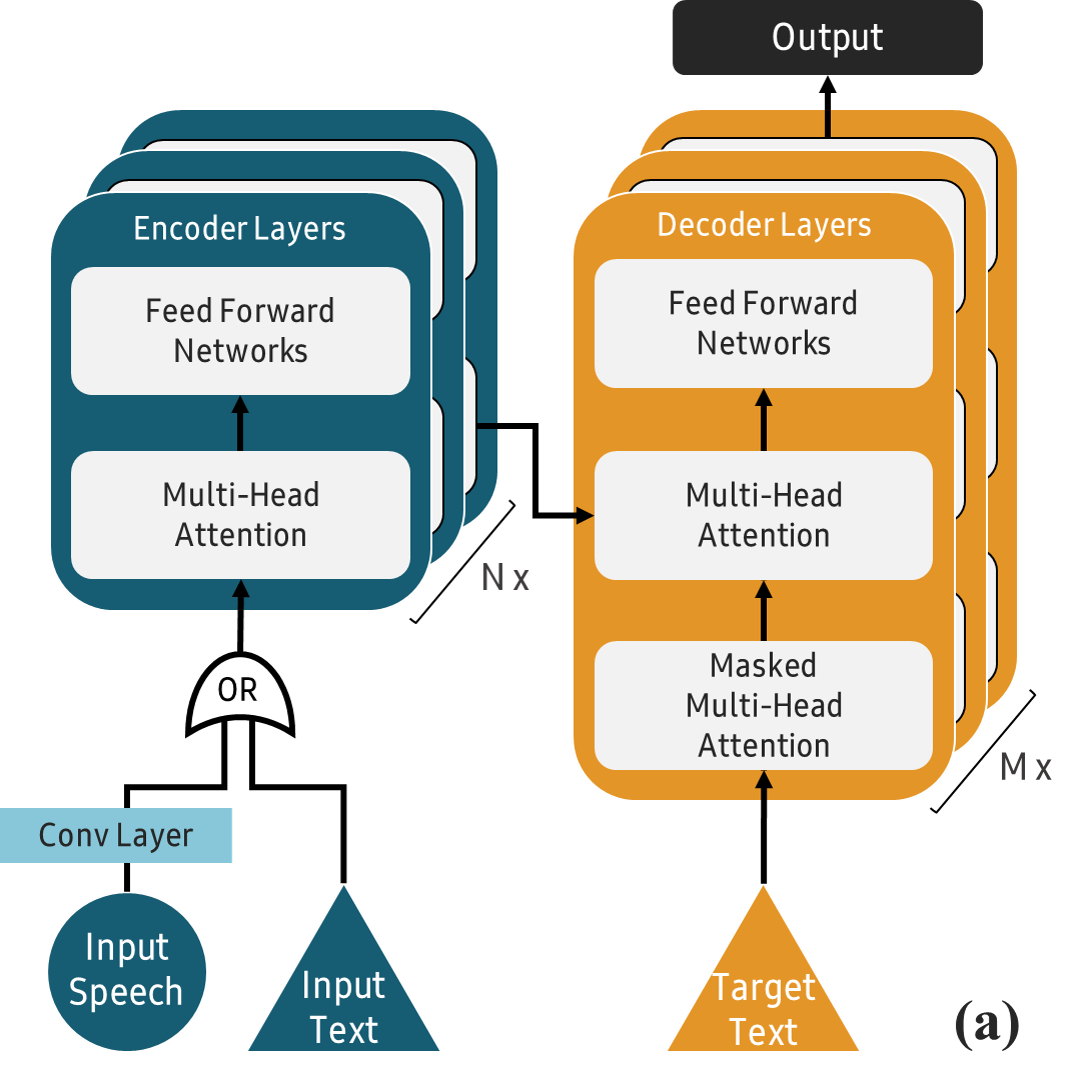}
        %\caption{}
        \label{fig:arch}
    \end{subfigure}%
    ~ 
    \hspace{0cm}
    \begin{subfigure}[t]{0.4\textwidth}
        \centering
        \includegraphics[height=6cm]{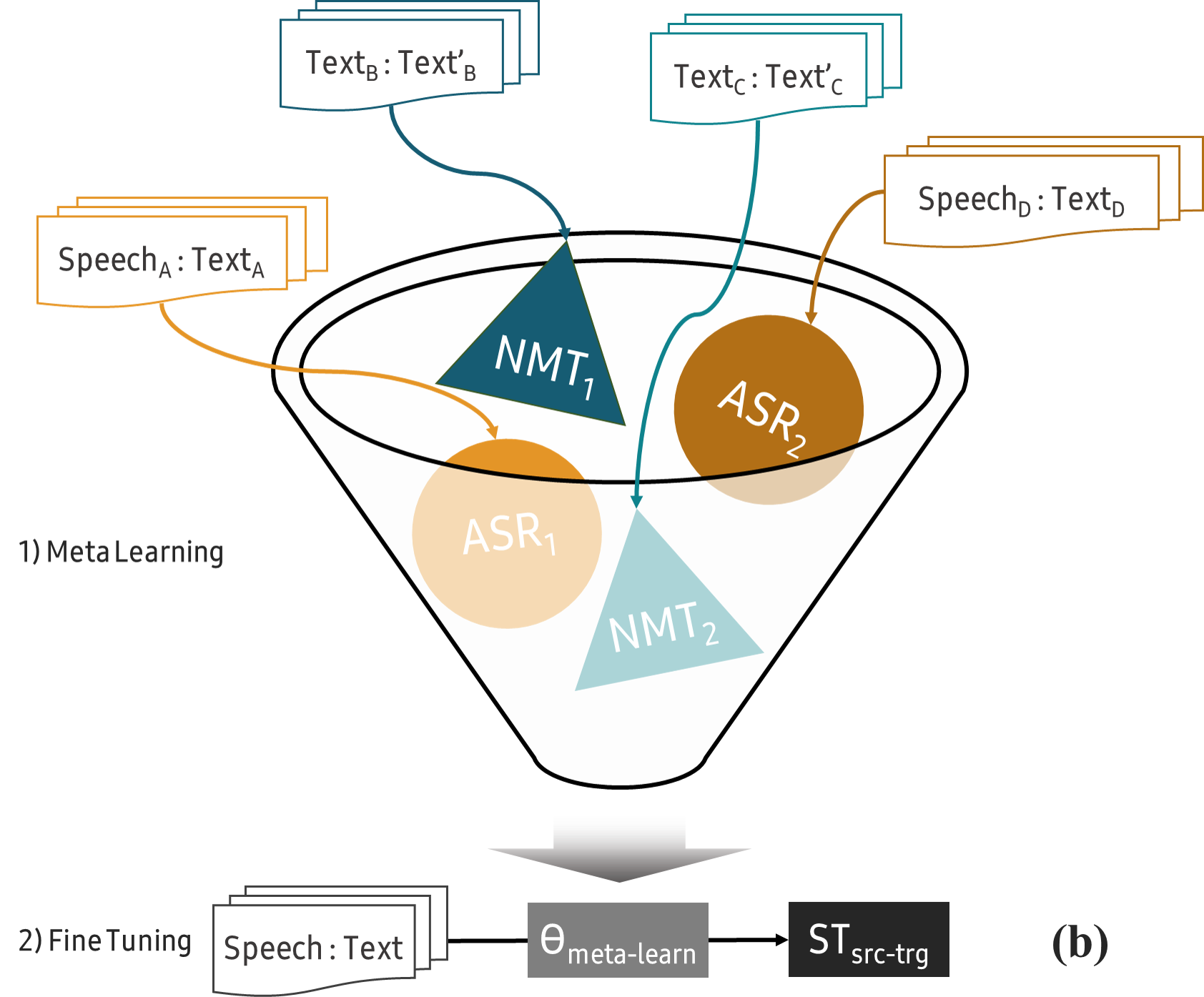}
        %\caption{}
        \label{fig:agent}
    \end{subfigure}
    \caption{\label{fig:overview}%
    (a) Overview of the base seq2seq architecture. (b) Overview of the modality agnostic meta learning model for ST.}
\end{figure*}
 
 To overcome the limitations of transfer-learning and its variants, we propose a multi-task learning approach based on a meta-learning algorithm, for the ST task. We adopt the model-agnostic meta-learning algorithm (MAML) \cite{finn2017model} to train on tasks with different input modalities. We use ASR and MT as source tasks during the meta-learning phase. These two tasks have different input modalities; ASR with speech input and MT with text input. The learned parameters from the meta-learning phase are used to initialize the parameters of our target ST model in the fine-tuning phase. There are two advantages of this approach over transfer learning: 1) The parameter updates of the model are not only based on the source ASR and MT tasks but also how good these works as initialization parameters for the target ST task. 2) We can utilize both the ASR and MT data at the same time without sharing parameters between auxiliary ASR, MT tasks, and the target ST task.

We conducted several experiments on English-German (En-De) and English-French (En-Fr) speech translation tasks from the MuST-C corpus \cite{mustc} to evaluate the effectiveness of the proposed meta-learning approach.  Our experiments reveal that the proposed approach achieves state-of-the-art performance on En-De, En-Fr ST tasks by obtaining 22.11 and 34.05 BLEU points, respectively.

\section{Speech Translation with Meta-Learning}
\subsection{Problem Formalization}
A typical Sequence-Sequence (seq2seq) architecture \cite{seq2seq} generates a target sequence $\boldsymbol{y}=\{y_1, \cdots, y_n\}$ given a source sequence $\boldsymbol{x} = \{x_1, \cdots, x_m\}$ by modeling the conditional probability,  $p(\boldsymbol{y}|\boldsymbol{x}, \theta)$. In general, the seq2seq architecture consists of two components: (1) an encoder which computes a set of representations $\boldsymbol{\Tilde{X}} = \{\boldsymbol{\Tilde{x}_1}, \cdots, \boldsymbol{\Tilde{x}_m}\} \in \mathbb{R}^{m \times d}$ corresponding to $\boldsymbol{x}$, and a decoder  coupled with attention mechanism \cite{Bahdanau14} dynamically reads encoder's output and predicts the distribution of each token in the target language. It is trained on a dataset of $D$ parallel sequences to maximize the log likelihood:

\begin{equation}
\label{eq:loss}
    \ell(D;\theta) = -\frac{1}{|D|}\sum_{i=1}^{N} \log p \left(\boldsymbol{y}^{i}|\boldsymbol{x}^{i};\theta \right),
\end{equation}
where $\theta$ are parameteres of the model.

The ASR, MT, and ST tasks in our work share the same seq2seq architecture.  The ASR and ST tasks take speech signal as input, and the input to the MT task is a sequence of characters or wordpiece tokens.
%where each speech signal is represented by 80 channel log mel filterbank features extracted from 25ms windows with a hop size of 10ms, stacked with delta and delta-delta features. 
The output of all the models is a sequence of tokens consisting of either characters or wordpiece tokens.

\subsection{Base Seq2Seq Model}
In recent years, the non-recurrent Transformer network \citet{vaswani2017attention} has achieved best translation quality for MT task. The encoder and decoder blocks of the Transformer are composed of a stack N and M identical layers, respectively. Each layer in the encoder contains two sublayers, a multi-head self-attention mechanism and a position-wise fully connected feed-forward network. Each decoder layer consists of three sublayers; the first and third sub-layers are similar to the encoder sub-layers, and the additional second sub-layer is used to compute the encoder-decoder attention (context) vector based on the soft-attention based approaches \cite{Bahdanau14}. Please refer to \citet{vaswani2017attention} for detailed architecture.

% The core module of the Transformer is the self-attention mechanism that weights different positions of input sequence to compute the representations for the inputs. It has three inputs: queries, keys and values. Each query output is computed as a weighted sum of the values, where each weight of the value is computed by a scaled dot product between query and the corresponding key. Let $Q \in \mathbb{R}^{t_{q} X d_{q}}$ be th queries, $K \in \mathbb{R}^{t_{k} X d_{k}}$ be the keys and $V \in \mathbb{R}^{t_v X d_v }$ be the values, where $t_*$ represents position of elements in the sequence and $d_*$ represents the dimension of elements. Generally, $t_k=t_v$ and $d_q = d_k$. The output of the self-attention are computed as follows:
% \begin{equation}
%     \label{eq:sa}
%     Attention (Q, k, V) = \mathrm{softmax} \left(\frac{QK^T}{\sqrt{d_k}}\right)V,
% \end{equation}
% where the scalar $1/\sqrt{d_k}$ is used to prevent softmax function into regions that have very small gradients. Please refer \citet{vaswani2017attention} for detailed architecture.

Here we adopt the above transformer network as our base seq2seq model for our ASR, MT, and ST tasks. Specifically, we adapt it to ASR and ST tasks by appending a compression layer. The speech sequence,  represented using Mel bank features, is commonly a few times longer than the text sequence. Therefore, we stack $3\times3$ CNN layers with stride 2 for both time and frequency dimensions to compress the length and exploit the structure locality of the speech signal. This compressed signal is later sent to the self-attention layers of the encoder. The overview of the base seq2se model is shown in Figure 1(a).

\begin{algorithm}[t]
\SetAlgoLined
\caption{Meta-Learning Algorithm for ST task}
\label{alg:maml}
% \SetAlgoLined
\textbf{Input}: Training examples from source tasks, $T=\{ASR, MT\}$ and target ST task. 

\textbf{Input}: Hyperparameters such as learning rates, $\alpha$ and $\beta$

Randomly initialize model parameters $\theta^m$.

\While{not done}{
  Sample task, $\tau$ from $T$
  
  Assign $\theta^a = \theta^m$
  
  Sample K data points, $D_\tau=\{x_{(i)}, y_{(i)}\}_{i=1}^{k}$ from $\tau$
  
  compute $\nabla_{\theta^{m}}\ell(D_\tau;\theta^{m})$ using $D_\tau$ and $\theta^m$
  
  \textbf{Meta-Train:} update $\theta^a$ using Eq. \ref{eq:aux-grad}
  
  sample l data points, $D_{\tau}^{'}=\{x_{(i)}^{'}, y_{(i)}^{'}\}_{i=1}^{l}$ from $\tau$
  
  compute $\nabla_{\theta^{a}}\ell(D_\tau^{'};\theta^{a})$ using $D_\tau^{'}$ and $\theta^m$
  
  \textbf{Meta-Test:} update $\theta^m$ using Eq. \ref{eq:meta-grad}
}

Assign $\theta=\theta^m$

\While{not done}{
  sample m data points, $D_{st} = \{x_{(i)}, y_{(i)}\}_{i=1}^{m} \in $ ST task
  
  compute $\nabla_{\theta} \ell(D_{st};\theta)$ using $D_{st}$ and $\theta$
  
  \textbf{Finetune:} Update $\theta$ with gradient descent: $\theta=\theta-\gamma \nabla_{\theta} \ell(D_{st};\theta)$
  
}

Return: $\theta$
 
\end{algorithm}

\subsection{Modality Agnostic Meta-Leaning for ST}
\label{sec:maml}
The base seq2seq model described above is known to easily overfit and result in an inferior performance when the training data is limited. We mitigate this issue by sharing the knowledge between low and high resource tasks using the MAML algorithm. The approach of  MAML is to use a set of high resource tasks as source tasks $\{\tau^1, \cdots, \tau^s\}$ to find a good parameter initialization point $\theta^0$ for the low resource target task $\tau^0$. 

\textbf{Meta-Learning Phase:} In this paper, we extend the idea of MAML to meta-learn on tasks with different input modalities. The source tasks in our model are ASR and MT with speech-text and text-text modalities, respectively. Later, we fine-tune the target ST task from the parameters of the meta-learned model ($\theta^{m}$). The overview of the proposed approach is shown in Figure 1(b). The process can be understood as
\begin{equation}
    \theta^* = \mathrm{Learn(ST; Meta-Learn(ASR, MT))}.
\end{equation}

We find the initialization $\theta^0$ for ST task by simulating low resource scenarios using source ASR and MT tasks. We define the meta objective function $ \ell(\theta^{m})$ to get $\theta^0=\theta^m$ following \citet{finn2017model}:
\begin{equation}
\label{eq:meta-obj}
    \ell(\theta^{m}) =E_\tau E_{D_k, D_{k}^{'}} \left[ \ell\left(D_{\tau};\ell\left(D_{\tau}^{'};\theta^{m}\right)\right)\right],
\end{equation}
where $\tau$ refers to the randomly sampled task to carry-out one meta-learning step. The set of samples $D_{\tau}$ and $D_{\tau}^{'}$ follow the uniform distribution over $\tau$'s dataset.

We maximize the meta-objective function in eq. \ref{eq:meta-obj} using gradient descent. For each meta-learning step, we uniformly sample one source task ($\tau$) at random from the set, $\{\mathrm{ASR, MT}\}$. We then sample two batches of training examples, $D_{\tau}$ and $D_{\tau}^{'}$, independently from the chosen source task, $\tau$. We use $D_\tau$ to simulate task-specific learning and the $D_{\tau}^{'}$ to evaluate its outcome. We call the gradient step to simulate task-specific learning (the  \textit{auxiliary-gradient} step). The auxiliary parameters ($\theta^{a}$) are updated using the  auxiliary-gradient step with the learning parameter $\alpha$, which is given as:
\begin{equation}
\label{eq:aux-grad}
    \theta_{\tau}^{a} = \theta^{m}-\alpha \nabla_{\theta^{m}}\ell(D_\tau;\theta^{m}).
\end{equation}

Once the task-specific learning is done, we evaluate the auxiliary parameters $\theta^{a}$ against the previously sampled batch of training examples, $D_{\tau}^{'}$. The gradient computed on ($\ell(D_{\tau}^{'};\theta^{a})$) during this evaluation is called the \textit{meta-gradient}. The meta parameters ($\theta^{m}$) are updated using this meta-gradient and is computed as follows:
\begin{equation}
\label{eq:meta-grad}
    \theta_{\tau}^{m} = \theta^{m}-\beta \nabla_{\theta^{a}}\ell(D_\tau^{'};\theta^{a}),
\end{equation}
where $\beta$ is the learning rate. Use of second derivates when estimating the meta-gradient through the auxiliary gradient in eq. \ref{eq:meta-obj} requires expensive Hessian matrix computation. Therefore, by following the vanilla MAML algorithm,  we also use first-order approximation while computing the meta-gradients.

The meta-learned parameters $\theta^{m}$, updated through eq. \ref{eq:meta-grad}, can adapt to a new learning task using only a small number of training examples.
% write about second order gradient

\textbf{Dealing with Different Modalities}: The vanilla MAML algorithm does not handle tasks with different input-output modalities. Moreover, we use additional compression layer on the input speech signal and it is not required for input text sequence. To deal with these limitations: (1) We create a universal vocabulary from all the tasks by following \citet{universalvocab}. (2) We dynamically disable the compression layer whenever we sample from the MT task during the meta-learning phase. That is, the MT examples do not affect the parameters of the compression layer. 

\begin{table*}[t]
\centering
\begin{tabular}{|c|c|c|c|c|c|}
%\hline 
\hline
Task & Dataset & Domain & Train & Dev & Test \\ \hline
\multirow{2}{*}{ST}& MuST-C English-German  & \multirow{2}{*}{TED Talks}  & 229K & 1.4K & 2.6K \\  
                   & MuST-C English-French  &                             & 275K & 1.4K & 2.6K \\ 
                  \hline  
ASR & Spoken Wikipedia Corpus-English & Wikipedia Articles &  347K & 2.7K & 2.0K  \\
\hline  
\multirow{2}{*}{MT} & WMT16 English-German & News articles \&& 4.5M & 3K & 3K \\ 
                    & WMT16 English-French & European Parliment proceedings   & 40.8M & 3K & 3K \\
                    \hline
\end{tabular}
\caption{Statistics of the datasets used in our experiments.}
\label{Table:datasets}
\end{table*}

\textbf{Fine-tuning Phase:}
 During the meta-learning phase, the parameters of the model ($\theta^{m}$) are exposed to vast amounts of speech-to-transcripts and text-to-text translation datasets via ASR and MT tasks. This allows the parameters of all the sublayers in the model such as compression, encoder, decoder, encoder-decoder attention, and output layers to learn individual language representations and translation relations between them. Hence, the meta-learned parameters ($\theta^{m}$) may not be suitable for the ST task on its own but can act as a good starting point to learn the target ST task. The model parameters are initialized from $\theta^{m}$ and further updated based on the target ST task evaluations. During fine-tuning phase, model training proceeds like in usual neural network training without involving auxiliary updates. An overview of the proposed modality agnostic meta-learning approach is given in Algorithm \ref{alg:maml}.

\section{Experiments}

% General Description of experiments 
\subsection{Datasets and Metrics}
\textbf{Target Tasks:} We used the MuST-C corpus \cite{mustc} to test the effectiveness of our proposed approach. MuST-C is a corpus for ST from English to 8 different target languages (German, Spanish, French, Italian, Dutch, Portuguese, Romanian and Russian). This corpus is created from English TED talks, which are automatically aligned at the sentence level with corresponding English transcriptions and translations to the target languages mentioned above. This dataset is larger than any other publicly available ST corpus.
%We exploit the presence of the (English) transcripts for augmenting the data with synthetic translations 

In our experiments, we focus on two target languages, German and French.  The En-De corpus consists of around 408 hours of English speech, which corresponds to 234k sentences of paired speech-transcript-translation data, whereas the En-Fr corpus has 492 hours of speech, corresponding to 280k sentences, see Table \ref{Table:datasets}. 

\noindent \textbf{Source Tasks: } We use the Spoken Wikipedia Corpus (SWC, \citet{swc})  for training the English ASR tasks. The SWC is a collection of time-aligned spoken Wikipedia articles for Dutch, English, and German using a fully automated pipeline to download, normalize and align the data.
% Being recorded by volunteers reading complete articles, this corpus represents natural speakers very well and is usually better than controlled recording in a lab.
We use the English speech and English transcripts of the SWC Corpus to train our ASR models in transfer learning and meta-learning phase. This corpus contains  352k sentences in 395 hours of speech read by a diverse set of 413 speakers. 
%This dataset is one of the largely known English ASR datasets.
%ASR performance is measured in terms of Word Error Rate(WER). 

For training MT tasks during meta-learning, we use the WMT16 En-De and En-Fr language pairs. The datasets are created by extracting language pairs from news articles and proceedings of the European Parliament. 

The datasets used in ST, ASR, and MT tasks come from different domains. The datasets of ST are from TED talks, ASR from Wikipedia and MT from news articles. The primary reason to use datasets from different domains is that it is difficult to gather all the task-specific datasets from the same domain. Therefore, here we test the generalization performance of ST task trained with the help of ASR and MT tasks collected from different domains. The statistics of all the datasets used in our experiments are shown in Table \ref{Table:datasets}.

\noindent \textbf{Data Processing and Evaluation Metrics:} The speech signal in ASR and ST is represented by log Mel 80-dimensional features. The text sequence in all the tasks is split into characters preserving word boundaries. We report the case sensitive BLEU scores on test sets for ST and MT tasks and are obtained using 4-gram NIST BLEU score \cite{Papineni:2002:BMA:1073083.1073135}. ASR performance is measured in terms of word error rate (WER). We choose the best models based on the dev set performance and report the results on the testset.

\subsection{Implementation Details}
The proposed model is implemented based on Tensor2Tensor framework \cite{tensor2tensor}. The number of convolutional layers in the compression layer is set to 2.  We use eight encoder and decoder layers in our experiments.  We apply dropout rate of 0.2 to the output of each sublayer before it is added to the sublayer input and normalized. 
%We also apply dropout rate of 0.1 to the sums of the embeddings and positional embeddings.
We use a batch size of 1.5M frames for ASR and ST tasks and a batch size of 4096 tokens for MT task.  All other hyperparameters such as optimization algorithm, learning rate schedule are set similar to \citet{vaswani2017attention}. All the models are trained on 4*NVIDIA V100 GPUs. 

% \begin{table*}[t]

% \centering

% \begin{tabular}{ |c|c|c|c|c|c|}
%  \hline
%  \multirow{2}{*}{S.No.} & \multirow{2}{*}{Model} & \multirow{2}{*}{Char / Word Piece} & Augment with & \multicolumn{2}{c|}{BLEU} \\
%                         &                        & & Synthetic Data    & En-De &    En-Fr  \\  \hline \hline

%  \multirow{2}{*}{1} & Transfer Learning [] & \multirow{2}{*}{char} & \multirow{2}{*}{No} & \multirow{2}{*}{12.93} & \multirow{2}{*}{22.29} \\ 
%                     &  (Baseline 1)        &  &                     &                     & \\ \hline
% \multicolumn{6}{c}{\textbf{This Work}} \\ \hline   
% %  &           &                         &    &       &       \\
% 2 & Transfer Learning & char & No & 15.60 & 26.94 \\
% %   &  &      &    &       &       \\
% 3 & (Baseline 2)      & char & Yes & 17.76 & 28.17 \\
%   %&                   &      &     &       &       \\
%   \hline \hline
%   &                   &      &     &       &       \\
% 4 & \multirow{5}{*}{} & char & No  & 17.20 & 29.19 \\
%   & &  &    &       &       \\
% \textbf{5} &   \textbf{Meta Learning}       & char & \textbf{Yes} & \textbf{20.02} & \textbf{31.09} \\
%           &          &            &              &                 &  \\ 
% \textbf{6} &          & \textbf{word piece} & \textbf{Yes} & \textbf{22.11} & \textbf{34.05} \\ \hline
% \end{tabular}

% \caption{Performance of various models on En-De and En-Fr speech translation tasks.}
% \label{Table:mainresult}
% \end{table*}

\begin{table*}[t]

\centering

\begin{tabular}{ |c|c|c|c|c|c|}
 \hline
 \multirow{2}{*}{No.} & \multirow{2}{*}{Model} & Char / & Synthetic & \multicolumn{2}{c|}{ST (BLEU)} \\
                        &                        & Wordpiece & Data Augmentation   & En-De &    En-Fr  \\  \hline \hline

 \multirow{2}{*}{1} & Transfer Learning \cite{mustc} & \multirow{2}{*}{char} & \multirow{2}{*}{No} &  \multirow{2}{*}{12.93} & \multirow{2}{*}{22.29} \\ 
                    &  (\texttt{Baseline 1})        &  &                &                     & \\ \hline
\multicolumn{6}{c}{\textbf{This Work}} \\ \hline   
2 & Transfer Learning (\texttt{Baseline 2}) & char & No & 15.60 & 26.94 \\
3 & Multi-Task Learning (\texttt{Baseline 3}) & char & No & 16.00 & 26.20 \\ 
\textbf{4} &   \textbf{Meta-Learning}   & \textbf{char} & \textbf{No} & \textbf{17.20} & \textbf{29.19} \\
\hline
5 & Cascade (\texttt{Baselien 4}) & wordpiece & Yes & 20.86 & 33.7 \\
\textbf{6} &    \textbf{Meta-Learning}      & \textbf{wordpiece} & \textbf{Yes} & \textbf{22.11} & \textbf{34.05} \\ \hline

\end{tabular}

\caption{Performance of various models on En-De and En-Fr speech translation tasks.}
\label{Table:mainresult}
\end{table*}

%% Add a main table for ende, and enfr. And mention about WMT, ASR data and how it is

%% Write about the paper numbers (and how they get their result?)
\subsection{Baselines}
We present the reported results from \citet{mustc} as \texttt{Baseline 1} to compare with our models. The architecture of \texttt{Baseline 1} is based on \cite{berard2018end} and it is an attention-based end-to-end ST model. The encoder of the model is based on feedforward, convolutional layers, and three stacked LSTMs. The decoder consists of a two-layered deep transition LSTM \cite{deeplstm}. The system is trained using transfer learning strategy by first pre-training the ASR model followed by the ST model. The ASR model is trained on  speech-to-transcripts available from the MuST-C corpus. 

In order to show the effectiveness of the proposed Meta-Learning (ML) approach compared to Transfer Learning (TL) and Multi-Task Learning (MTL) approaches, we design two baselines, called as \texttt{Baseline 2 and Baseline 3}. The architecture of these baselines is precisely similar to the seq2seq model used in our meta-learning experiments. These baselines are more powerful compared to \texttt{Baseline 1} and act as better baselines to compare the effectiveness of the proposed meta-learning approach. The \texttt{Baseline 2}  is pre-trained on the ASR task, and \texttt{Baseline 3} is pre-trained on all the three tasks (ASR, MT, and  ST) simultaneously. The two baselines are further fine-tuned on the ST task. The datasets used during the pretraining phase in these approaches are same as the meta-learning phase.

% Our best model 
\subsection{Main Results} 
\textbf{Baseline 1 vs. Baseline 2: }
We compare our non-recurrent based \texttt{Baseline 2} model (in Table \ref{Table:mainresult}, model no. 2) against the \texttt{Baseline 1}. Even though the ASR dataset used during transfer learning of our \textit{Baseline 2} is out of domain with ST task, it achieved significant BLEU score improvement for both En-De ($\uparrow 2.67$) and En-Fr ($\uparrow 4.65$) ST tasks compared to the \textit{Baseline 1}. Therefore, we use \textit{Baseline 2} to compare our proposed meta-learning approach.

\noindent \textbf{Meta-Learning vs. Transfer Learning: } 
Here, we compare the performance of \texttt{Baseline 2, Baseline 3}, and the proposed modality agnostic meta-learning model for ST task. During the meta-learning phase, we use the SWC English dataset for the ASR task and WMT dataset of English to the corresponding ST task target language for the MT task. The parameters of the model are updated using the proposed meta-learning approach, as described in Section \ref{sec:maml}. 
% Here we compare the performance of our \textit{Baseline 2} and the proposed modality agnostic meta-learning model for ST task. During the meta-learning phase, we use SWC English dataset for the ASR task and WMT dataset of English to the corresponding ST task target language for the MT task. The parameters of the  model are updated using the proposed meta-learning approach, as described in Section \ref{sec:maml}. We train the meta-learn model for 800K steps on both the SWC, WMT dataset combined. For each training step during this phase, one task is randomly sampled, and then the model is trained on that task, so each task roughly gets trained for 400k steps. The transfer learning based \textit{Baseline 2} is also trained for 400K steps on SWC English corpus. Fine-tuning phase of  the ST model is started by initializing the parameters either from the transfer or meta-learned model and carried out for 600k steps on the MuST-C Corpus.
From Table \ref{Table:mainresult}, we can see that the ST model trained using the proposed meta-learning approach outperforms \texttt{Baseline 2 and Baseline 3} by achieving 17.20  and 29.19 BLEU score on En-De and En-Fr language pairs, respectively.  The results show that the meta-learning phase helps to learn the individual language representations and relations between them. Moreover, we can see that the meta-learning algorithm helps the target task despite being trained on the source tasks coming from different domains.

\begin{table}[h]
\centering
\begin{tabular}{|c|c|c|c|}
%\hline 
\hline
\multirow{2}{*}{ST Task} & TL & \multicolumn{2}{c|}{ML} \\ 
                         &   ASR (wer $\downarrow$)             &  ASR (wer $\downarrow$) & MT (bleu $\uparrow$) \\ \hline
                En-De    &   37.43                 &  42.95     & 17.16 \\ 
                En-Fr    &   37.43                 &  39.56     & 24.70 \\ \hline
\end{tabular}
\caption{Performace of various source tasks used in transfer learning (TL) and meta-learning (ML) approaches.}
\label{Table:auxresult}
\end{table}

We also report the performance of ASR, MT tasks used in transer and meta-learning phase in Table \ref{Table:auxresult}. The performance of the ASR model used in transfer learning approach is significantly better than the ASR model in meta-learning approach. However, we achieved significantly better results for target ST task with the meta-learning approach. This is expected given that in the meta-learning phase, we update the parameters with a focus to adapt to the target ST task instead of focusing heavily on learning the particular source task. This also applies to the MT model, whose performance is lower than the standard MT model.

\begin{figure*}[t]
   \centering
    \begin{subfigure}[t]{0.5\textwidth}
       \centering
        \includegraphics[height=5cm]{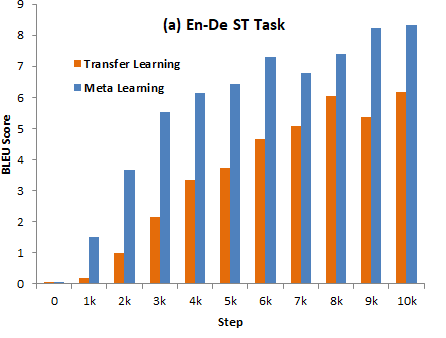}
        %\caption{En-De ST Task}
        \label{fig:edblue}
    \end{subfigure}%
    ~ 
    \begin{subfigure}[t]{0.5\textwidth}
        \centering
        \includegraphics[height=5cm]{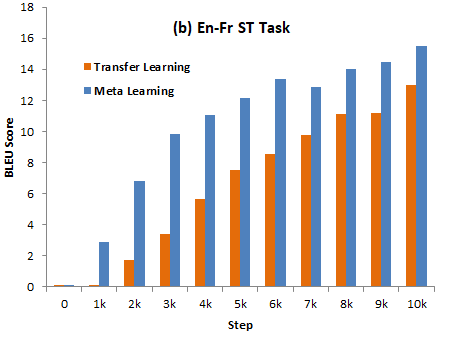}
        %\caption{En-Fr ST Task}
        \label{fig:enfrbleu}
    \end{subfigure}
    \caption{\label{fig:overview}%
    ST model performance on testset obtained from the checkpoints 0 to 10k.
    %(a) Training loss over En-De ST task. (b) Training loss over En-Fr ST Task.
    }

\end{figure*}

\begin{figure*}[t]
   \centering
    \begin{subfigure}[t]{0.5\textwidth}
       \centering
        \includegraphics[height=5cm]{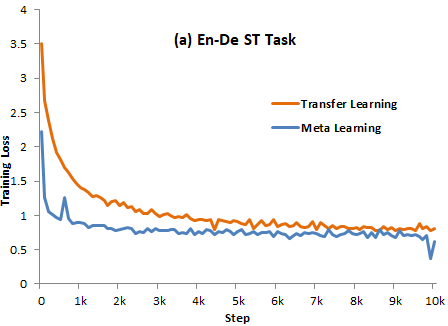}
        %\caption{En-De ST Task}
        \label{fig:edloss}
    \end{subfigure}%
    ~ 
    \begin{subfigure}[t]{0.5\textwidth}
        \centering
        \includegraphics[height=5cm]{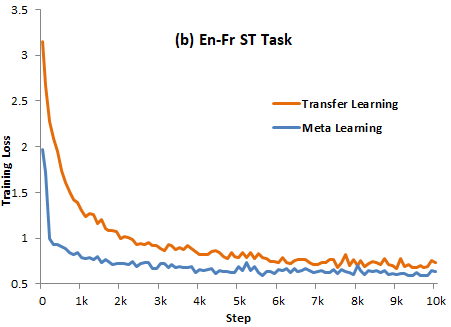}
        %\caption{En-Fr ST Task}
        \label{fig:enfrloss}
    \end{subfigure}
    \caption{\label{fig:overview}%
    ST model loss from the training steps 0 to 10k.
    %(a) Training loss over En-De ST task. (b) Training loss over En-Fr ST Task.
    }
\end{figure*}

\subsection{Impact of Initialization}
To study the effectiveness of the meta-learned parameters ($\theta^{m}$) as an initialization point and check quick adaptability to the target ST task, we analyze the BLEU scores and training losses for the first few steps. We compare the models obtained by fine-tuning from the meta-learned parameters ($\theta^{m}$) against the transfer learning parameters ($\theta^{t}$). We measure the BLEU score for every 1K steps for the first 10K steps for both the models on the test set. We can see from Figures 2(a) and 2(b) that the meta learned parameters act as a much better initialization point for the target task. We also plot the training loss on the ST task for the first 10k steps from the models obtained by fine-tuning from ($\theta^{m}$) vs. ($\theta^{t}$). From Figures 3(a) and 3(b),  we can observe that the ST model initialized from the meta-learned parameters has significantly lesser loss than that from the transfer learned parameters. Here, we reported only the first 10K steps of BLEU score and training loss to see the effect near the initialization step. However, these trends continued in the later training steps as well. 

\subsection{Sample Translations}
We present a few sample translations from the testset in Table \ref{Table:samples} for the transfer and meta-learning apporaches. These are generated using model numbers 2 and 4 in Table \ref{Table:mainresult}. Analyzing these samples gives an insight into the proposed meta-learning approach for the ST task. We can see that the translations from the meta-learning approach preserve the context better than the ones from the transfer learning approach. We also see that it mitigates the speech pronunciation issues by leveraging language representations learnt via MT task during the meta-learn phase (For example, Confucianism versus Confisherism in the last example in Table \ref{Table:samples}).

\subsection{Further Improvements}
We further improved the results of our approach by (1) Augmenting the MuST-C corpus ST data using synthetic data \cite{jia2019leveraging}. We first train the MT model using the Transformer network on the WMT16 MT dataset and later generate new translations using MuST-C transcripts. We combine the generated translation and original speech signal to create the synthetic training point. (2) Training wordpiece \cite{subwords} vocabulary based models, instead of character-based models. We adopt \cite{universalvocab} to create a universal vocabulary based on all the tasks present in our meta-learning approach.

Our meta-learning approach with these additional improvements achieves new state-of-the-art results on En-De and En-Fr ST tasks by obtaining a BLEU score of 22.11 and 34.05, respectively, and surpasses cascaded system (\texttt{Baseline 4}). The ASR model in the cascaded system is trained on the SWC corpus and fine-tuned on MuST-C transcripts, and the MT model is pre-trained on WMT16 and fine-tuned on Must-C transcript-translation data.

\begin{table*}[t]
\centering
\begin{tabular}{|l|p{16.4cm}|}
%\hline 
\hline
TS & \textbf{(En)} So the same as we saw before. \\
OT  & \textbf{(De)} Also genauso, wie wir es vorher gesehen haben. \textbf{(BT)} Just as we have seen before. \\
TL & \textbf{(De)} Das Gleiche gilt f{\"u}r uns. \textbf{(BT)} The same \textcolor{red}{applies to} us. \\ 
ML & \textbf{(De)} Das Gleiche haben wir also schon fr{\"u}her gesehen. \textbf{(BT)} So we saw the same thing earlier.\\ \hline 
TS & \textbf{(En)} This is what, in engineering terms, you would call a real time control system. \\
OT & \textbf{(De)} Dies w{\"u}rden Sie, in Ingenieurteams, eine Echtzeit-Kontrollsystem nennen. \textbf{(BT)} You would call this, in engineering teams, a real-time control system. \\
TL & \textbf{(De)} Das ist es, was man im Ingenieurssystem nennen k{\"o}nnte. \textbf{(BT)} That's what you could call in the engineering system. \\
ML & \textbf{(De)} Das ist es, was man in Ingenieurwissenschaften als Echtzeit-Kontrollsystem bezeichnen k{\"o}nnte. \textbf{(BT)} That's what you could call a \textcolor{blue}{real-time control system} in engineering. \\
\hline \hline
TS & \textbf{(En)} They even can bring with them some financing. \\
OT & \textbf{(Fr)} Ils peuvent m{\^e}me apporter avec eux des financements. \textbf{(BT)} They can even bring with them funding. \\
TL & \textbf{(Fr)} Ils peuvent m{\^e}me les amener avec eux et financer. \textbf{(BT)} They can even bring \textcolor{red}{them with them} and finance. \\
ML & \textbf{(Fr)} Ils peuvent m{\^e}me apporter avec eux des financements. \textbf{(BT)} They can even bring with them funding. \\
      \hline
TS & \textbf{(En)} She grew up at a time when Confucianism was the social norm and the local mandarin was the person who mattered. \\
OT & \textbf{(Fr)} Elle a grandi {\`a} une époque o{\`u} le confucianisme {\'e}tait la norme sociale et le mandarin local {\'e}tait la personne qui importait. \textbf{(BT)} She grew up at a time when Confucianism was the social norm and local Mandarin was the person who mattered. \\
TL & \textbf{(Fr)} Elle a grandi {\`a} un moment o{\`u} le confisherisme {\'e}tait le norme social, et la mandarine locale {\'e}tait la personne qu'il avait import{\'e}e. \textbf{(BT)} She grew up at a time when \textcolor{red}{Confisherism} was the social norm, and the local mandarin was the person he had imported. \\
ML & \textbf{(Fr)} Elle a grandi {\`a} une {\'e}poque o{\`u} le confucianisme {\'e}tait la norme sociale, et la mandarine locale {\'e}tait la personne qui comptait. \textbf{(BT)} She grew up in a time when \textcolor{blue}{Confucianism} was the social norm, and the local mandarin was the person who counted. \\
      \hline
      
\end{tabular}
\caption{Sample translation from transfer and meta-learning (TL, ML) approches  for En-De and En-Fr ST tasks. We provided transcripts (TS), original translations (OT), and back translations (BT) from De/Fr$\rightarrow$En to help the the readers.}
\label{Table:samples}
\end{table*}

\section{Related Work}
\label{sec:Related work}

\noindent \textbf{End-to-End Speech Translation: }
Traditionally, speech translation is implemented as a cascade  of ASR and  MT \cite{ney1999speech,cascade1}.  However, it has its own limitations. 
%The cascaded model not only requires high computational resources and computing time but also suffers from compounding error, which limits the performance of the whole system. 
Starting with the attempt to align source speech and target translation text without transcription \cite{duong2016attentional,anastasopoulos2016unsupervised}, several works have been proposed to realize the end-to-end speech translation system \cite{berard2016listen,weiss2017sequence,berard2018end}. 
%However, training such end-to-end systems requires a substantial amount of speech-text translation pairs to avoid overfitting, and 
However, collecting such huge speech-to-translation corpus is relatively more challenging compared to the collection of MT and ASR corpora. 
This challenge leads to attempting various methods to moderate its paucity, including data augmentation with MT or TTS (Text-to-Speech) models, and utilization of data of other related tasks by employing transfer learning.  
Augmenting training data with synthesized audio using TTS is also adopted \cite{berard2016listen,kano2018structured,jia2019leveraging}. %
Several variants of transfer learning approach such as multi-task learning have been explored by simultaneously training either ASR+ST or MT+ST pairs \cite{weiss2017sequence,multistmt}. 
However, the performance gains were similar to the transfer learning approach \cite{berard2018end}. The above approaches result in sub-optimal solutions for the target ST task due to the reasons discussed in Section \ref{sec:intro}. 
%In contrast to the previous approaches, we approach the multi-task learning in such a way that the parameters of ST task are not shared with any other auxiliary tasks and at the same time we utilize the data from related high resource tasks. 
%low resource \cite{besacier2006towards}
%low resource \cite{besacier2006towards}

\noindent \textbf{Meta-Learning: }
In general, meta-learning, or learning-to-learn, aims to solve the problem of adapting to new tasks with few examples. The meta-learning focuses more on learning aspects instead of wholly focusing on a particular task at hand. Several approaches have been proposed for meta-learning to acquire an ability of fast adaptation. \citet{bengio1992optimization,andrychowicz2016learning,ha2016hypernetworks} approach the meta-learning by learning a meta-policy, while  \citet{finn2017model,vinyals2016matching} learn to find a good initialization point for a new task. 
% For few shot learning, several approaches are proved to be effective by comparing unseen examples in a trained metric space with use of siamese network \cite{koch2015siamese} or recurrent with attention mechanisms \cite{santoro2016meta}.
Our work is based on the later approaches, specifically, it is based on the recent  model agnostic meta-learning (MAML, \citet{finn2017model}) that can be readily applied to any gradient descent based neural network.
% Inspired by MAML, \cite{gu2018meta} suggest an approach to mitigate the problem of low resource MT by setting different language translation as a separate task and exploiting high resource language pair for target low resource language pair with meta-learning.
% check with HJ
%One of the methods is to learn updating rules 
Our work is similar in spirit to the work of low resource neural machine translation \cite{gu2018meta}. However, we focus on adapting meta-learning to tasks with different input modalities and solve the more challenging ST task. 
%\citet{maml++} recently proposed efficient techniques to train MAML, we left that to our future work.

\section{Conclusion}

In this work, we introduce a modality agnostic meta-learning to solve the low resource end-to-end speech translation task. The proposed approach adopts from MAML and extends it to work on tasks with different modalities during the meta-learning phase. Our approach has several benefits. It makes use of vast amounts of data available from MT and ASR tasks and does not share parameters across the source and target tasks. It finds a good initialization point during the meta-learning using the source tasks=ASR+MT and adapts quickly to the target ST task during the fine-tuning phase. To test the effectiveness of the proposed approach, we conducted several experiments on En-De and En-Fr ST tasks. Our approach significantly outperforms the existing approaches of transfer learning on both the ST tasks. We further improved the performance of the proposed method by augmented synthetic data and using wordpiece vocabularies. 

The proposed approach brings new opportunities to build efficient end-to-end ST systems with a limited amount of training data. First, the approach incorporates ASR and MT tasks in a principled way to leverage additional sources of data. Second, it is a generic framework that can comfortably accommodate existing and future end-to-end ST models.
%In the future, we investigate the new approaches of meta-learning and effiecint architectures suitable across all the tasks involved in the meta-learning experminet to further improve the speech translation performance.

\section{Acknowledgments}
We thank Yoonsuck Choe for valuable discussions and comments.
\bibliography{aaai20}
\bibliographystyle{aaai}

\end{document}